\documentclass{article} 
\usepackage{iclr2025_conference,times}

\usepackage{amsmath,amsfonts,bm}

\def\eqref#1{equation~\ref{#1}}

\def\1{\bm{1}}

\DeclareMathAlphabet{\mathsfit}{\encodingdefault}{\sfdefault}{m}{sl}
\SetMathAlphabet{\mathsfit}{bold}{\encodingdefault}{\sfdefault}{bx}{n}

\usepackage{hyperref}
\usepackage{url}
\usepackage{graphicx}
\usepackage{multirow}
\usepackage{xspace}
\usepackage{booktabs} 

\makeatletter
\DeclareRobustCommand\onedot{\futurelet\@let@token\@onedot}
\def\@onedot{\ifx\@let@token.\else.\null\fi\xspace}

\makeatother

\usepackage{tcolorbox}
\tcbuselibrary{breakable}
\newtcbox{\textbox}[1][red]
  {on line, arc = 0pt, outer arc = 0pt,
    colback = #1!10!white, colframe = #1!50!black,
    boxsep = 0pt, left = 1pt, right = 1pt, top = 2pt, bottom = 2pt,
    fontupper=\ttfamily\bfseries\upshape,
    boxrule = 0pt, bottomrule = 1pt, toprule = 1pt}

\title{Visual-O1: Understanding Ambiguous Instructions via Multi-modal Multi-turn Chain-of-thoughts Reasoning}

\author{Minheng Ni$^{1,2}$, Yutao Fan$^2$, Lei Zhang$^1$ \& Wangmeng Zuo$^2$\\
$^1$Department of Computing, Hong Kong Polytechnic University \\
\texttt{minheng.ni@connect.polyu.hk} \quad \texttt{cslzhang@comp.polyu.edu.hk} \\
$^2$Faculty of Computing, Harbin Institute of Technology\\
\texttt{fanyutao@stu.hit.edu.cn} \quad \texttt{wmzuo@hit.edu.cn}\\
}

\iclrfinalcopy 
\begin{document}

\maketitle

\begin{abstract}
As large-scale models evolve, language instructions are increasingly utilized in multi-modal tasks. Due to human language habits, these instructions often contain ambiguities in real-world scenarios, necessitating the integration of visual context or common sense for accurate interpretation. However, even highly intelligent large models exhibit significant performance limitations on ambiguous instructions, where weak reasoning abilities of disambiguation can lead to catastrophic errors. To address this issue, this paper proposes \textsc{Visual-O1}, a multi-modal multi-turn chain-of-thought reasoning framework. It simulates human multi-modal multi-turn reasoning, providing instantial experience for highly intelligent models or empirical experience for generally intelligent models to understand ambiguous instructions. Unlike traditional methods that require models to possess high intelligence to understand long texts or perform lengthy complex reasoning, our framework does not significantly increase computational overhead and is more general and effective, even for generally intelligent models. Experiments show that our method not only significantly enhances the performance of models of different intelligence levels on ambiguous instructions but also improves their performance on general datasets. Our work highlights the potential of artificial intelligence to work like humans in real-world scenarios with uncertainty and ambiguity. We will release our data and code.
\end{abstract}

\section{Introduction}

With the advancement of deep learning, increasing attention is being paid to multi-modal scenarios that are more relevant to reality, such as visual question answering (VQA) \citep{antol2015vqa,zhou2020unified,teney2018tips}, referring image segmentation (RIS) \citep{lai2023lisa,ni2023ref,yang2023set}, and vision-and-language navigation (VLN) \citep{li2023improving,hong2021vln,feng2022uln}. In recent years, more manipulable language instructions are gradually being introduced into these tasks to better align with human interaction habits. Combined with large-scale language models \citep{GPT-4o,chen2023minigpt,liu2024improved}, artificial intelligence (AI) is beginning to use language instructions closer to real-world scenarios to perform tasks, significantly expanding the scope of AI applications.

\begin{figure}[h]
	\centering
	\includegraphics[width=14cm]{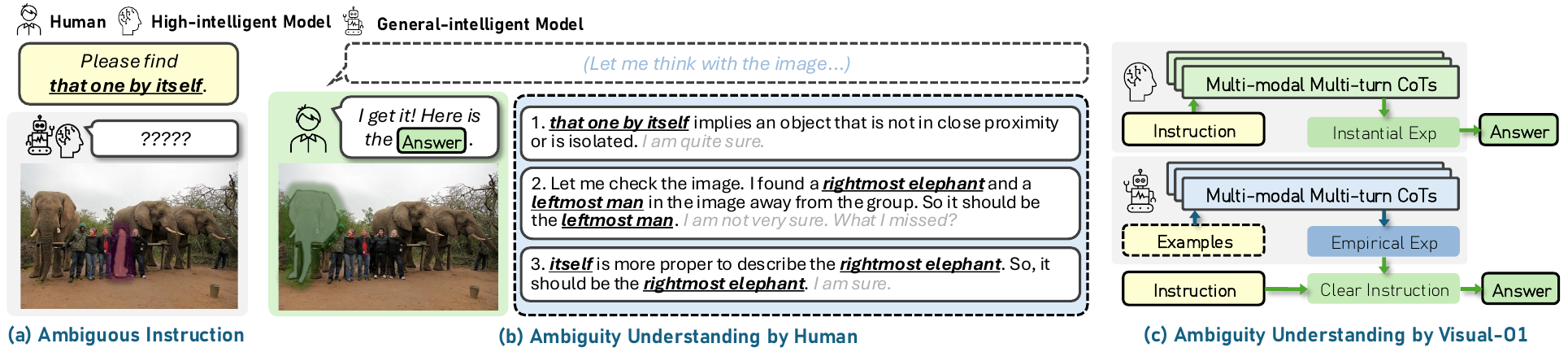}
	\caption{\textbf{Understanding ambiguous instruction.} The AI model may not be able to execute instructions normally when encountering ambiguous instructions. However, humans can usually correctly analyze the actual meaning of ambiguous instructions by combining visual context and can accurately interpret ambiguous instructions. Based on this, we propose \textsc{Visual-O1}, which simulates human multi-modal multi-turn reasoning to gain instantial (for high-intelligent models) or empirical (for general-intelligent models) experience in order to understand ambiguous instructions.}
	\label{fig:intro}
\end{figure}

As shown in Figure \ref{fig:intro}, due to the inherent ambiguity of natural language and the excellent analytical abilities of humans, the language instructions used by humans often contain vagueness and ambiguity. Additionally, human language and vision are closely related and often require combining visual context or common sense to accurately understand their meanings. Therefore, ambiguous instructions, which are common in reality, differ from meticulously designed accurate instructions \citep{antol2015vqa}, and learning to understand them directly becomes challenging in the absence of corresponding task data.

Recently, chain-of-thoughts (CoT) reasoning has significantly enhanced the understanding and analytical capabilities of high-intelligent large models, such as \textsc{GPT-4o}. However, its application to scenarios involving multi-modal ambiguity understanding has yet to be explored. Additionally, general-intelligence models used in multi-modal tasks, like \textsc{LLaVA}, often lack the data and parameter capacity to perform chain reasoning and analysis, making it difficult to apply CoT methods to enhance their understanding of ambiguous instructions.

To address these challenges, we propose \textsc{Visual-O1}, a multi-modal multi-turn chain-of-thought reasoning method that simulates human multi-modal multi-turn reasoning. \textsc{Visual-O1} builds instance-specific experience during inference for high-intelligence models or creates general experience for any ambiguous instructions through one-time optimization with several examples for general-intelligence models. This helps models correctly understand ambiguous instructions and synthesize the final answer.

We also construct a dataset containing various types of ambiguous instructions, including \textit{ellipsis}, \textit{colloquialism}, \textit{subjectivity}, \textit{relativity}, and \textit{other}, to validate performance across different multi-modal scenarios. Experiments show that our method significantly improves the performance of models with varying intelligence levels on the ambiguous instruction dataset and enhances their performance on general datasets. Ablation studies demonstrate that \textsc{Visual-O1} can be easily applied to different multi-modal models and tasks.

Our contributions are three-fold:
\begin{itemize}
    \item We reveal the capabilities of multi-modal models in analyzing and executing ambiguous instructions by setting up a novel benchmark for understanding ambiguous instructions in various multi-modal tasks.
    \item We propose \textsc{Visual-O1}, a multi-modal multi-turn chain-of-thought reasoning method, to build instantial or empirical experience for high-intelligent or general-intelligent models, enabling them to correctly understand ambiguous instructions.
    \item Experimental results show that our method significantly improves the performance of models with varying intelligence levels on ambiguous instruction datasets and enhances their performance on general datasets.
\end{itemize}
\section{Related Work}

\paragraph{Language Instruction Understanding in Multi-modal Tasks}

Language instructions were often used as conditions in multi-modal tasks, serving as an essential medium for user-AI interaction. They had been applied in a wide range of highly reality-related tasks \citep{reed2016generative,bigham2010vizwiz,zellers2019recognition}. Traditionally, language instructions were precise and unambiguous. Although some works had managed to understand and execute complex language instructions by combining large multi-modal models (LMMs) \citep{lai2023lisa, yang2023set}, they still lacked a practical understanding of ambiguous instructions.
Recently, a few works had noticed the presence of ambiguity in language within specific multi-modal tasks.
In image classification, \textsc{WaffleCLIP} \citep{roth2023waffling} and \textsc{FuDD} \citep{esfandiarpoor2023follow} had pointed out the issue of polysemy in classification texts, using LLMs to generate commonalities or differences within categories to enhance image classification tasks.
In visual question answering, \citet{prasad2023rephrase} proposed \textsc{Rephrase} to repeatedly utilize LLMs to mine image information through manually pre-designed prompts to enhance language understanding.

However, existing methods required models to engage in extensive interactions with samples based on predefined rules to mine information, requiring extremely long inference times and relying on specific tasks and models, making it difficult to generalize to various tasks and models with different intelligence levels.

\paragraph{Complex Reasoning with Large Multi-modal Models}

In recent years, large language models (LLMs) have expanded into multi-modal scenarios \citep{GPT-4o, liu2024improved, lai2023lisa}, showcasing the potential of large multi-modal models (LMMs) in handling multi-modal tasks. However, challenges remain in understanding complex language instructions within visual contexts in multi-modal tasks. Recently, methods to enhance the reasoning capabilities of large multi-modal models have attracted some research attention. On one hand, some works have proposed training-based methods.
\citet{dai2024instructblip} proposed a two-stage training method that aligned pre-trained models with images and texts, enhancing the capabilities of LMMs. Similarly, \citet{chen2023sharegpt4v, bai2023qwen, wang2024qwen2} achieved impressive results in multi-modal tasks using data generated by GPT. On the other hand, some works have proposed non-training methods, such as the chain-of-thoughts (CoT) \citep{wei2022chain,yao2024tree,yao2022react} approach, which enhances model understanding by simulating human reasoning.

However, existing methods either relied on a large amount of real or model-generated data to optimize model parameters, which is challenging to scale cost-effectively to any task, or they required the model to have strong reasoning abilities, making it challenging to general-intelligent models.
\section{Visual-O1: Multi-modal Multi-turn Chain-of-thoughts Reasoning Framework}

\begin{figure}[h]
	\centering
	\includegraphics[width=14cm]{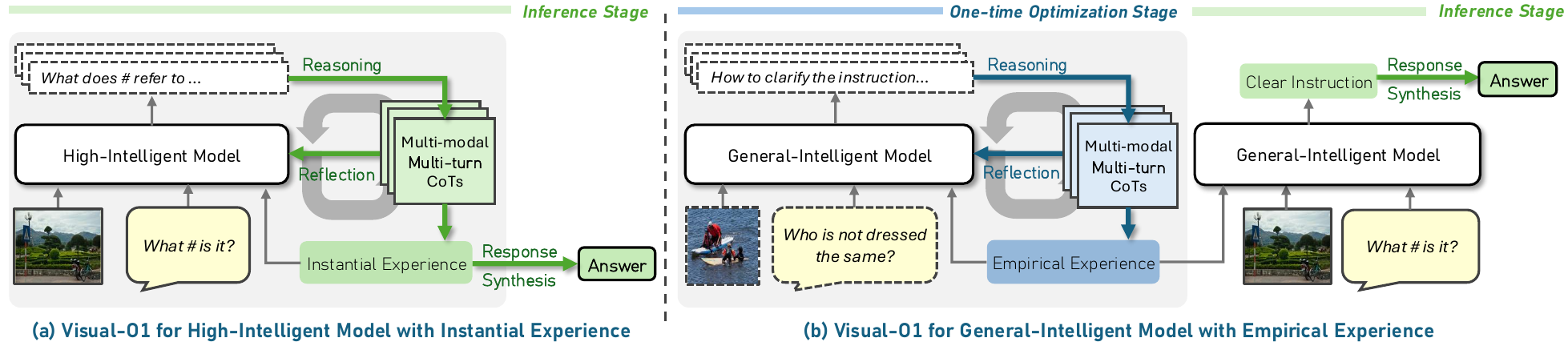}
	\caption{\textbf{The overview of \textsc{Visual-O1}.} \textsc{Visual-O1} introduces multi-modal multi-turn chain-of-thoughts to understand ambiguity with (a) instantial experience for high-intelligent models to generate the correct answer directly, and (b) empirical experience for general-intelligent models to transform ambiguous instructions into clear instructions and then generate the correct answer. Instantial and empirical experience develops during inference and one-time optimization stage.}
	\label{fig:method}
\end{figure}

\subsection{Overview of Proposed Framework}

Given a task that requires both a visual context $\mathbf{x}_v$ and an ambiguous instruction $\mathbf{x}_a$, it can be performed by a multi-modal model $f$. The result of the task is denoted as $\mathbf{y} = f(\mathbf{x}_a, \mathbf{x}_v \mid \theta_f)$, where $\theta_f$ represents the parameters of the task model\footnote{Since we do not optimize $\theta_f$, we omit it in the following text for brevity}. Due to the ambiguous or unclear information contained in the text instruction $\mathbf{x}_a$, the performance of the model $f$ can be significantly affected. If this model is a general-intelligent model, this can have catastrophic effects due to its weaker reasoning abilities, making it nearly impossible for the model to produce the correct result. Even for high-intelligent large multi-modal models, we still observe significant performance limitations (see Tables \ref{tab:overall_ris} and \ref{tab:overall_vqa} for more details).

To address this issue, as shown in Figure \ref{fig:method}, we propose \textsc{Visual-O1}, a multi-modal multi-turn chain-of-thoughts reasoning framework. \textsc{Visual-O1} aims to help models build experience in disambiguating ambiguous instructions, enabling them to correctly understand the instructions and produce the right answers. Unlike traditional chain-of-thought methods that require models to possess high intelligence for understanding long texts and performing complex reasoning, our framework is more general and effective even for general-intelligent models.

Specifically, for high-intelligent models, \textsc{Visual-O1} establishes sample-level disambiguation experience during inference by leveraging the ambiguous instructions themselves. For general-intelligent models, \textsc{Visual-O1} builds empirical disambiguation experience during one-time optimization using several examples. \textsc{Visual-O1} synthesizes responses by either summarizing the sample-level experience into the final answer or using the empirical experience to transform ambiguous instructions into clear instructions. This clear instruction is then combined with the original ambiguous instruction to summarize the final answer. Despite the slight differences in Visual-O1 for high-intelligent and general-intelligent models, their structures are completely consistent. Therefore, we can formulate them within a unified framework.

\subsection{Reasoning and Reflection}

Given a multi-modal model $f$, by utilizing the visual context and logical reasoning, we can supplement vague information, enabling the model to understand the precise meaning of the ambiguous instruction. To achieve this, we perform a reasoning step under the guidance of a pre-defined prompt $p_{\mathrm{rsn}}$ to attempt to understand the ambiguous instruction:
\begin{align}
\mathbf{x}_{\mathrm{rsn}} &= f(\mathbf{x}_a, \mathbf{x}_v \mid p_{\mathrm{rsn}}; \mathcal{A}),
\end{align}
where $\mathbf{x}_{\mathrm{rsn}}$ is the reasoning result, and $\mathcal{A}$ is a special prompt that contains the disambiguation experience. Our goal is to obtain this special prompt $\mathcal{A}$ to help $f$ understand the ambiguities.

Assuming we complete one reasoning step, next, the model reflects on the reasoning result $\mathbf{x}_{\mathrm{rsn}}$. Assuming $p_{\mathrm{rfl}}$ is the predefined prompt, we obtain the reflection result $\mathbf{x}_{\mathrm{rfl}}$ by:
\begin{align}
\mathbf{x}_{\mathrm{rfl}} &= f(\mathcal{A}, \mathbf{x}_{\mathrm{rsn}} \mid p_{\mathrm{rfl}}).
\end{align}

By utilizing reflection, we summarize the issues in the reasoning process to improve the disambiguation experience $\mathcal{A}$. By repeatedly performing the reasoning-reflection process, we gradually refine the reasoning results, transforming it into an iterative form. We rewrite the reasoning and reflection steps in an iterative form as follows:
\begin{align}
\mathbf{x}_{\mathrm{rsn}}^{(i)} &= f(\mathbf{x}_a, \mathbf{x}_v \mid p_{\mathrm{rsn}}; \mathcal{A}^{(i)}),\label{eq:3}\\
\mathbf{x}_{\mathrm{rfl}}^{(i)} &= f(\mathcal{A}^{(i)}, \mathbf{x}_{\mathrm{rsn}}^{(i)} \mid p_{\mathrm{rfl}}),\label{eq:4}
\end{align}
where $i$ is the $i$-th iteration. Assuming our budget is $N$, we directly apply the iterations formed by Eq. (\ref{eq:3}-\ref{eq:4}) and an update function of $\mathcal{A}^{(i)}$ by $N$ times to obtain the final $\mathcal{A}:=\mathcal{A}^{(N)}$. Next, we introduce the instantial and empirical experience of $\mathcal{A}$ and their updating mechanism in Visual-O1 for different intelligence levels of the model.

\paragraph{Reasoning and Reflection for Instantial Experience}
For high-intelligent models, in each iteration, \textsc{Visual-O1} continues to analyze the ambiguous instruction based on the previous disambiguation process, then reflects on its analysis and generates feedback. These analyses and feedback form the reasoning process for the entire problem. Therefore, for the $i$-th iteration, we update $\mathcal{A}_\mathrm{ins}$ by aggregating all information:
\begin{align}
\mathcal{A}^{(i+1)}_{\mathrm{ins}} &= \mathcal{A}^{(i)}_{\mathrm{ins}} \oplus \mathbf{x}_{\mathrm{rsn}}^{(i)} \oplus \mathbf{x}_{\mathrm{rfl}}^{(i)},\label{eq:5}
\end{align}
where $\oplus$ is the concatenation function. Obviously, this process can directly work on the inference stage. Assuming our budget is $N_{\mathrm{ins}}$, we directly apply the iterations formed by Eq. (\ref{eq:3}-\ref{eq:5}) $N_{\mathrm{ins}}$ times during the inference stage to obtain a complete disambiguation process. 

\paragraph{Reasoning and Reflection for Empirical Experience}
For general-intelligent models, in each iteration, \textsc{Visual-O1} uses the existing experience to continue attempting to transform ambiguous instructions into clear ones, then reflects on its conversion process and updates the experience. Due to the limited understanding capabilities of general intelligent models for long texts, we only retain the new reflection information as the update for $\mathcal{A}_\mathrm{emp}$:
\begin{align}
\mathcal{A}^{(i+1)}_{\mathrm{emp}} &= \mathbf{x}_{\mathrm{rsn}}^{(i)}.\label{eq:6}
\end{align}

Similar to the process of a high-intelligent model, we apply the iterations formed by Eq. (\ref{eq:3}-\ref{eq:4}, \ref{eq:6}) $N_{\mathrm{emp}}$ times to obtain the final disambiguation experience. The difference is that we change one sample in each iteration, which means a total of $N_{\mathrm{emp}}$ samples, to prevent a learning experience that is specific to certain samples rather than a generalizable experience. Fortunately, we only need to obtain $\mathcal{A}_{\mathrm{emp}}$ once in the one-stage optimization stage; for the same task, we no longer need to repeatedly obtain $\mathcal{A}_{\mathrm{emp}}$.

\subsection{Response Synthesis}

Although experience $\mathcal{A}$ is not the final answer, it contains the model's understanding of the instructions and is important information for solving the problem. Next, we obtain the final answer $\mathbf{y}$ through response synthesis.

\paragraph{Response Synthesis by Instantial Experience}
For high-intelligent models, $\mathcal{A}_{\mathrm{ins}}$ already encompasses all the reasoning, and the model possesses strong comprehension abilities. Therefore, we simply summarize the entire reasoning process with a pre-defined response synthesis prompt $ p_{\mathrm{syn}}$ to directly obtain the final answer:
\begin{equation}
\mathbf{y}=f(\mathcal{A}_{\mathrm{ins}}\mid p_{\mathrm{syn}}),
\label{eq:x}
\end{equation}

\paragraph{Response Synthesis by Empirical Experience}
For general-intelligent models, considering their limited reasoning capabilities, we first use the disambiguation experience $\mathcal{A}_{\mathrm{emp}}$ learned by the model during the one-time optimization process to transform the ambiguous instruction $\mathbf{x}_a$ into a clear instruction $\mathbf{x}_c$ based on the visual context $\mathbf{x}_v$:
\begin{equation}
\mathbf{x}_c = f(\mathbf{x}_a, \mathbf{x}_v \mid \mathcal{A}_{\mathrm{emp}}).
\end{equation}

Then, we utilize the pre-defined response synthesis prompt $p_{\mathrm{syn}}$, allowing the model to summarize the original ambiguous instruction and the clear instruction together to obtain the final answer:
\begin{equation}
\mathbf{y} = f(\mathbf{x}_c, \mathbf{x}_a, \mathbf{x}_v \mid p_{\mathrm{syn}}). \label{eq:x2}
\end{equation}

Finally, regardless of the model's level of intelligence, we obtain a final answer $\mathbf{y}$ that well-understands the ambiguous instruction $\mathbf{x}_a$ based on visual context $\mathbf{x}_v$.

\subsection{Implementation Details}

We fix the random seed as $42$ for all experiments. We set budgets $N_{\mathrm{ins}}$ and $N_{\mathrm{emp}}$ to $10$ and $3$ for general-intelligent and high-intelligent models respectively. 
Please note that due to the difficulty of performing reflection with general-intelligent models, in practice, we temporarily use a high-intelligent model at this step. Since the reflection for general-intelligent models is only conducted during a one-time optimization, our final inference still relies solely on the general-intelligent model itself without introducing additional models or computational overhead.

For more details of implementation, please refer to the \textbf{Appendix} \ref{app:imp}.
\section{Experiments}

\subsection{Experimental Setup}

To verify the effectiveness of our method, we conduct a series of experiments. We apply \textsc{Visual-O1} to the state-of-the-art high-intelligence model and general-intelligence model on two typical multi-modal tasks: referring image segmentation (RIS) and visual question answering (VQA). All comparisons are divided into ambiguous instructions and general instructions to comprehensively evaluate the models' performance on ambiguous and non-ambiguous instructions.

\paragraph{Baselines and Evaluation Metrics.} We select a series of typical methods as baselines. We choose the original, unprocessed high-intelligence \textsc{SoM}~\citep{yang2023set} and \textsc{GPT-4o}~\citep{GPT-4o} as well as the general-intelligence LISA~\citep{lai2023lisa} and \textsc{LLaVA}~\citep{liu2024improved} as comparison models for RIS and VQA. Then, we apply \textsc{Chain-of-thoughts}~\citep{wei2022chain} reasoning and \textsc{FuDD}~\citep{esfandiarpoor2023follow} language explanation methods on LISA and \textsc{LLaVA} for further comparisons. To ensure fairness in the comparisons, we ensure that all models are under the same settings. Following previous work \citep{bigham2010vizwiz, kazemzadeh2014referitgame, ni2023ref}, for RIS, we compare gIoU and cIoU, and for VQA, we compare accuracy and BLEU-1.

\paragraph{Datasets.} In addition to the general datasets, we set up extra ambiguous datasets for RIS and VQA, which only contain ambiguous instructions filtered or refined manually. We categorize the ambiguous instructions into five types based on the cause of ambiguity: \textit{ellipsis}, \textit{colloquialism}, \textit{subjectivity}, \textit{relativity}, and \textit{other}. \textit{Ellipsis} indicates that the ambiguity stems from omitted content; \textit{colloquialism} indicates that the ambiguity arises from the use of informal or imprecise expressions; \textit{subjectivity} indicates that the ambiguity is due to subjective judgments; \textit{relativity} indicates that the ambiguity comes from implied comparisons, and \textit{other} represents other types of ambiguity.

For more details of the dataset, please refer to the \textbf{Appendix} \ref{app:data}.

\subsection{Overall Results}
\label{sec:exp_overall}

\begin{table}[t!]
\caption{\textbf{Overall results on RIS.} \textsc{Visual-O1} significantly outperforms other methods and achieves notable improvements on both high-intelligent and general-intelligent models.}
\label{tab:overall_ris}
\begin{center}
\begin{small}
\begin{sc}
\begin{tabular}{lcccccccc}
\toprule
\multirow{2}{*}{Model} & \multicolumn{2}{c}{Ambiguous} & &  \multicolumn{2}{c}{General}\\
\cmidrule{2-3}\cmidrule{5-6} & gIoU & cIoU & & gIoU & cIoU \\
\midrule
LISA \citep{lai2023lisa} & $0.0237$ & $0.0272$ && $0.4654$ & $0.4721$\\
SoM \citep{yang2023set} & $0.0752$ & $0.1158$ && $0.3507$ & $0.4154$\\
Chain-of-thoughts \citep{wei2022chain} & $0.0746$ & $0.0982$  && $0.4183$ & $0.4326$ \\
FuDD \citep{esfandiarpoor2023follow} & $0.0718$ & $0.0928$ && $0.3416$ & $0.3534$ \\
Visual-O1 (LISA)  & $\mathbf{0.1088}$ & $\mathbf{0.1209}$ && $\mathbf{0.4794}$ & $\mathbf{0.5004}$\\
Visual-O1 (SoM)  & $\mathbf{0.1278}$ &$\mathbf{0.1722}$ && $\mathbf{0.3602}$ & $\mathbf{0.4566}$ \\
\bottomrule
\end{tabular}
\end{sc}
\end{small}
\end{center}
 \vskip -0.2in
\end{table}

\begin{table}[t!]
\caption{\textbf{Overall results on VQA.} \textsc{Visual-O1} significantly outperforms other methods and achieves notable improvements on both ambigous and general datasets.}
\label{tab:overall_vqa}
\begin{center}
\begin{small}
\begin{sc}
\begin{tabular}{lcccccccc}
\toprule
\multirow{2}{*}{Model} & \multicolumn{2}{c}{Ambiguous} & &  \multicolumn{2}{c}{General}\\
\cmidrule{2-3}\cmidrule{5-6} & Acc & BLEU-1 & & Acc & BLEU-1 \\
\midrule
LLaVa \citep{liu2024improved} & $8.58$ & $0.4760$ && $54.19$ & $0.7194$\\
GPT-4o \citep{GPT-4o} & $14.10$ & $0.5960$ && $61.25$ & $0.8192$\\
Chain-of-thoughts \citep{wei2022chain} & $15.03$ & $0.3372$  && $39.78$ & $0.5113$ \\
FuDD \citep{esfandiarpoor2023follow} & $19.18$ & $0.4844$ && $48.37$ & $0.6454$ \\
Visual-O1 (LLaVa) & $\mathbf{21.35}$ & $\mathbf{0.5123}$ && $\mathbf{57.38}$ & $\mathbf{0.7329}$\\
Visual-O1 (GPT-4o) & $\mathbf{22.20}$ &$\mathbf{0.6475}$ && $\mathbf{63.25}$ & $\mathbf{0.8339}$ \\
\bottomrule
\end{tabular}
\end{sc}
\end{small}
\end{center}
\end{table}

\paragraph{Ambiguous instructions understanding.} As shown in the left part of Tables \ref{tab:overall_ris} and \ref{tab:overall_vqa}, we observe that the original model performs poorly on the distinctly different tasks of VQA and RIS. This indicates that existing methods are susceptible to ambiguous instructions and do not perform well when correct visual context needs to be integrated to understand instructions. After integrating \textsc{Visual-O1}, the performance of all tasks significantly improves, surpassing $100\%$. This means that existing methods have the potential to understand ambiguous instructions but require proper guidance. It is worth mentioning that we observe that using \textsc{Visual-O1}, general-intelligent model, \textsc{LLaVA}, and LISA are capable of achieving results comparable to or even better than high-intelligent models without introducing any additional models or data during inference. This further demonstrates the value of \textsc{Visual-O1}. 

\paragraph{General instructions understanding.} While the ability to understand ambiguous instructions is crucial, the performance of the model on general datasets is also important. As shown in the right part of Tables \ref{tab:overall_ris} and \ref{tab:overall_vqa}, we also find that \textsc{Visual-O1} significantly enhances the understanding of ambiguous instructions and greatly improves performance on general datasets. This is because improving the ability to understand instructions is beneficial even for those that do not contain ambiguity. What is even more noteworthy is that traditional \textsc{CoT} and description-based synthesis like \textsc{FuDD} significantly degrade the model's performance on general datasets. This is because general-intelligent models like \textsc{LLaVA} find it challenging to maintain the logical chains and long-text comprehension akin to high-intelligent large models like \textsc{GPT-4o}. Complex reasoning tends to impair the model's understanding capabilities. 

For more analysis on improvement, please refer to the \textbf{Appendix} \ref{app:improve}.

\subsection{Generalizability Studies}

\subsubsection{Flexibility to Intelligence Levels}
\label{sec:exp_int}

Based on the strong long-text comprehension and reasoning abilities of \textsc{GPT-4v}'s \textsc{SoM}, we use \textsc{Visual-O1} for instantial experience during the inference stage for each sample. In contrast, due to LISA's weaker reasoning abilities, we utilize \textsc{Visual-O1} for empirical experience during the one-time optimization phase, avoiding complex reasoning for each sample during the inference stage. However, this does not mean that \textsc{Visual-O1} cannot perform empirical experience for high-intelligence models or instantial experience for general-intelligence models. To test \textsc{Visual-O1}'s adaptability to different intelligence levels, we design additional models: \textsc{Visual-O1}$^+$(LISA) using instantial experience and \textsc{Visual-O1}$^-$(\textsc{SoM}) using empirical experience.

\begin{table}[t!]
\caption{\textbf{Ambiguous instruction results on different intelligence level.} \textsc{Visual-O1} has demonstrated powerful flexibility across different levels of intelligence.}
\label{tab:int_ris_amb}
\begin{center}
\begin{small}
\begin{sc}
\begin{tabular}{lccc}
\toprule
Model & Visual-O1 & gIoU & cIoU \\
\midrule
LISA \citep{lai2023lisa} &  & $0.0237$ & $0.0272$ \\
Visual-O1 (LISA) & Empirical & $\mathbf{0.1088}$ & $\mathbf{0.1209}$ \\
Visual-O1$^+$ (LISA) & Instantial & $\mathbf{0.1447}$ & $\mathbf{0.1351}$ \\
SoM \citep{yang2023set} &  &  $0.0237$ & $0.0272$ \\
Visual-O1$^-$ (SoM) & Empirical & $\mathbf{0.1143}$ & $\mathbf{0.1530}$ \\
Visual-O1 (SoM) & Instantial & $\mathbf{0.1278}$ & $\mathbf{0.1722}$ \\
\bottomrule
\end{tabular}
\end{sc}
\end{small}
\end{center}
 \vskip -0.2in
\end{table}

As shown in Table \ref{tab:int_ris_amb}, the variants of \textsc{Visual-O1} are effective across different models. It is noteworthy that since LISA cannot comprehend and generate long texts in logical chains, we use \textsc{GPT-4v} as its disambiguation model during the inference stage to supplement its capabilities. Whether applied to high-intelligent models or general-intelligent models, the variants of Visual-O1 significantly enhance performance on ambiguous instructions.

For more analysis on intelligence levels, please refer to the \textbf{Appendix} \ref{app:int}.

\subsubsection{Adaptability to Models}
\label{sec:exp_model}

\begin{table}[t!]
\caption{\textbf{Generalization studies on different models.} \textsc{Visual-O1} is effective across different models. ``-" indicates values not reported in the original paper.}
\label{tab:model_ris}
\begin{center}
\begin{small}
\begin{sc}
\begin{tabular}{lcc}
\toprule
Model & gIoU & cIoU \\
\midrule
OVSeg \citep{liang2023open} & $0.0418$ & $0.0778$ \\
Ref-Diff \citep{ni2023ref} & $0.3301 $ & $0.3140$ \\
Unified-IO \citep{lu2022unified} & $-$ & $0.4015$ \\
InstructDiffusion \citep{geng2023instructdiffusion} & $-$ & $0.3904$ \\
SoM \citep{yang2023set} & $0.3507$ & $0.4154$ \\
LISA \citep{lai2023lisa} & $0.4654$ & $0.4721$ \\
GPT-4o \citep{GPT-4o} & $0.5728$ & $0.5152$ \\
Qwen-VL-2 \citep{wang2024qwen2} & $0.4417$ & $0.3220$ \\
\midrule
Visual-O1 (SoM) & $\mathbf{0.3602}$ & $\mathbf{0.4566}$ \\
Visual-O1 (LISA) & $\mathbf{0.4794}$ & $\mathbf{0.5004}$ \\
Visual-O1 (GPT-4o) & $\mathbf{0.5777}$ & $\mathbf{0.5325}$  \\
Visual-O1 (Qwen-VL-2) & $\mathbf{0.5161}$ & $\mathbf{0.4457}$ \\
\bottomrule
\end{tabular}
\end{sc}
\end{small}
\end{center}
 \vskip -0.2in
\end{table}

Can \textsc{Visual-O1} generalize to different models? We select various models on RIS to verify this, and as shown in Table \ref{tab:model_ris}, regardless of the model used, we observe significant improvements. This highlights the generalizability of \textsc{Visual-O1} to different models and suggests the potential for broader applications of our method.

For more analysis on adaptability to different models and tasks, please refer to the \textbf{Appendix} \ref{app:task}.

\subsection{Ablation Studies}

\subsubsection{Effectiveness of Components}

To further explore the effectiveness of \textsc{Visual-O1}, we conduct ablation experiments on the general VQA data in Table \ref{tab:abl}. Reasoning and reflection are crucial for the model to correctly understand ambiguous instructions. The model's performance on ambiguous data significantly decreases without reasoning and reflection. Meanwhile, response synthesis is very important for the performance on regular data from non-ambiguous instructions, as the original instruction may contain important information, and response synthesis ensures the complete transmission of this information. Every module of \textsc{Visual-O1} significantly improves the model's performance across different datasets.

\begin{table}[t!]
\caption{\textbf{Ablation studies.} Each module of \textsc{Visual-O1} plays an indispensable role.}
\label{tab:abl}
\begin{center}
\begin{small}
\begin{sc}
\begin{tabular}{lcc}
\toprule
Model & Acc & BLEU-1 \\
\midrule
Visual-O1 (LLaVA) & $\mathbf{57.38}$ & $\mathbf{0.7329}$ \\
\quad w/o Response Synthesis & $47.59$ & $0.6589$ \\
\quad w/o Reasoning and Reflection & $54.25$ & $0.7076$ \\
\midrule
Visual-O1 (GPT-4o) & $\mathbf{63.25}$ & $\mathbf{0.8339}$ \\
\quad w/o Response Synthesis & $60.00$ & $0.8254$ \\
\quad w/o Reasoning and Reflection & $60.60$ & $0.7943$ \\
\bottomrule
\end{tabular}
\end{sc}
\end{small}
\end{center}
 \vskip -0.2in
\end{table}

For more ablations on components, please refer to the \textbf{Appendix} \ref{app:abl}.

\subsubsection{Exploration of Reasoning}

We design additional validation experiments to further prove the effectiveness of \textsc{Visual-O1}'s reasoning process. We invite 10 volunteers to manually design disambiguation empirical experiences, which are used as benchmarks for ten tests and averaged. As shown in Table \ref{tab:turn}, manually designed prompts are significantly weaker than reasoned empirical experience, proving the rationality of \textsc{Visual-O1}'s design.

\begin{table}[t!]
\caption{\textbf{Efficiency of reasoning in \textsc{Visual-O1}.} Automatic reasoning surpasses manually designed experience, and an appropriate budget can help improve the performance of \textsc{Visual-O1}.}
\label{tab:turn}
\begin{center}
\begin{small}
\begin{sc}
\begin{tabular}{lcc}
\toprule
Method & Acc & BLEU-1 \\
\midrule
Visual-O1 (LLaVA) w/o Budget & $54.25$ & $0.7076$ \\
\quad w/ Human Experience & $54.99$ & $0.7073$ \\
\quad w/ 1 Budget & $55.87$ & $0.7190$ \\
\quad w/ 2 Budget & $54.16$ & $0.7049$ \\
\quad w/ 3 Budget & $\mathbf{57.38}$ & $\mathbf{0.7329}$ \\
\quad w/ 4 Budget & $55.27$ & $0.7147$ \\
\quad w/ 5 Budget & $55.56$ & $0.7167$ \\
\bottomrule
\end{tabular}
\end{sc}
\end{small}
\end{center}
\end{table}

We also show the influence of budgets in the reasoning and reflection process of \textsc{Visual-O1}. The score is the lowest without reasoning and reflection, even lower than with manually designed prompts. However, with the reasoning and reflection process, the performance gradually grows and then falls again after reaching its peak. We observe that \textsc{Visual-O1} follows a trend similar to deep learning. The performance begins to grow in the early stages of learning, but then decreases due to overfitting after reaching a particular stage.

For more explorations on reasoning, please refer to the \textbf{Appendix} \ref{app:process}.

\subsubsection{Comparison of Data Augmentation}

\begin{table}[t!]
\caption{\textbf{Comparison with data augmentation.} Compared to resource-intensive data augmentation, \textsc{Visual-O1} still achieves significant advantages.}
\label{tab:aug}
\begin{center}
\begin{small}
\begin{sc}
\begin{tabular}{lcccc}
\toprule
Method & Acc & BLEU-1 \\
\midrule
LLaVA w/ Ambiguous Extra Data & $51.48$ & $0.6863$ \\
LLaVA w/ Ambiguous Original Data & $45.08$ & $0.6363$ \\
LLaVA w/ Noised Original Data & $39.84$ & $0.5912$ \\
Visual-O1 (LLaVA) & $\mathbf{57.38}$ & $\mathbf{0.7329}$ \\
\bottomrule
\end{tabular}
\end{sc}
\end{small}
\end{center}
 \vskip -0.2in
\end{table}

We generate an additional 10,000 data samples for fine-tuning \textsc{LLaVA}, as shown in Table \ref{tab:aug}. We design three different data augmentation methods: (1) \textsc{Ambiguous Extra Data}: directly using \textsc{LLaVA} to synthesize an extra 10,000 sets of ambiguous instruction data; (2) \textsc{Ambiguous Original Data}: rewriting 10,000 sets of original instructions into ambiguous instructions, then training the model; and (3) \textsc{Noised Original Data}: randomly deleting or modifying 10,000 sets of original instructions, then training the model.

We find that due to the synthetic data's inability to effectively simulate instructions, there is a significant performance drop in VQA. Additionally, the annotation and training costs brought by the 10,000 data samples are also very high. Therefore, \textsc{Visual-O1} effectively solves the problem of understanding ambiguous instructions while maintaining performance on the general dataset.

For comparisons of computational cost, please refer to the \textbf{Appendix} \ref{app:comp}.

\subsection{Case Studies}

\begin{figure}[h]
	\centering
	\includegraphics[width=14cm]{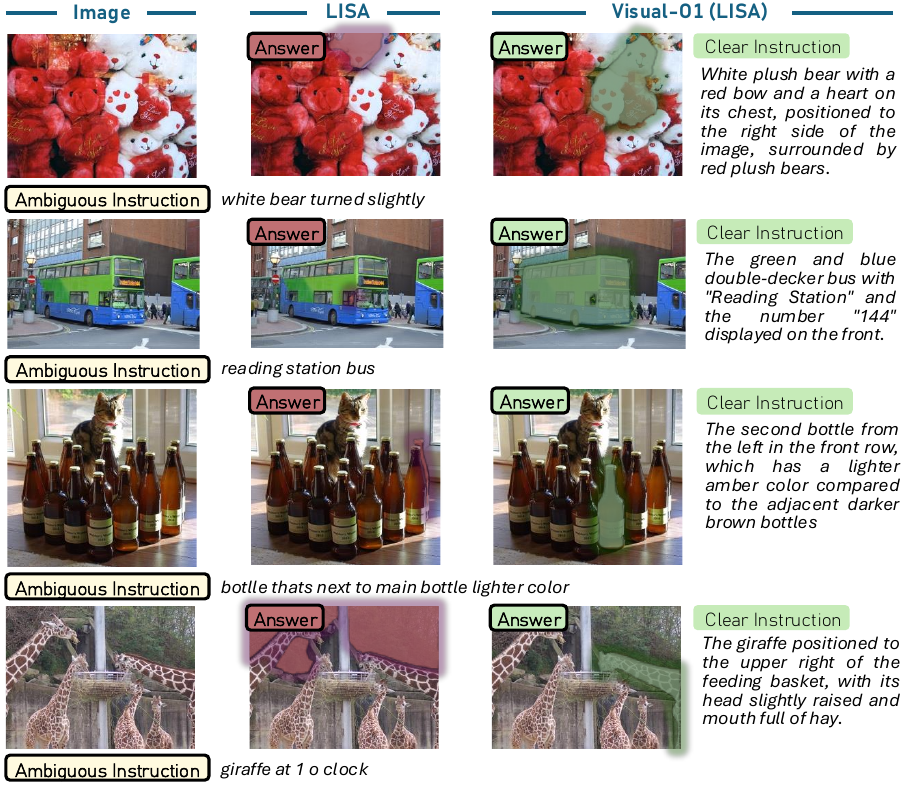}
	\caption{\textbf{Case studies on RIS.} Our approach aids the model in understanding ambiguous instructions by incorporating \textsc{Visual-O1}, which significantly improves the accuracy of instructions, thus enabling more effective segmentation of the target.}
	\label{fig:ris}
\end{figure}

How does our method help the model better understand ambiguous instructions? To delve into this issue, we visualize the results of RIS in Figure \ref{fig:ris}. We find that ambiguous instructions are common in the data and often not easily detected by humans, but this can severely affect the model's performance. After introducing \textsc{Visual-O1}, the accuracy of the instructions significantly improves, resulting in more effective outcomes.

As shown in the first example, this is a typical case of \textit{subjectivity} ambiguity, requiring the perspective of human subjective observation for reasonable inference. The original instruction could describe most of the bears in the image, as each bear's angle has some deviation. However, when combined with the image, it is found that only one bear meets the requirements of turning and slight deviation, and \textsc{Visual-O1} successfully infers this and provides an accurate description, enabling the task model to accurately segment the target.

In the second example, this is a typical case of \textit{ellipsis} ambiguity, where many sentence components are omitted. The segmentation model is misled by the word ``\texttt{reading}" and incorrectly segments the person in the image. Meanwhile, \textsc{Visual-O1} accurately describes that this is a bus showing a ``\texttt{reading station}" sign and even provides additional information to help the segmentation model locate the target.

In the third example, this is a typical case of \textit{relativity} ambiguity, where comparison is used to refer to specific entities. To understand the instruction, the model must first locate the comparison object of the bottle's lighter color, which is the color of other bottles, that is, the lightest-colored bottle. When combined with \textsc{Visual-O1}, the instruction is transformed into a more accurate form, enabling the model to easily locate the target.

In the fourth example, this is a typical case of \textit{colloquialism} ambiguity, using common positional phrases in spoken language. The model cannot understand that ``\texttt{1 o'clock}" is a position, so it cannot select the correct object, but \textsc{Visual-O1} transforms it into a clearer description.

For more cases, please refer to the \textbf{Appendix} \ref{app:case}.
\section{Conclusion}

Even high-intelligent large models exhibited significant performance limitations on ambiguous instructions, where weak reasoning abilities of disambiguation could lead to catastrophic errors. We proposed \textsc{Visual-O1}, a multi-modal multi-turn chain-of-thoughts reasoning framework. It simulated human multi-modal multi-turn reasoning, providing instantial experience for high-intelligent models or empirical experience for general-intelligent models to understand ambiguous instructions. Unlike traditional methods that required models to possess high intelligence to understand long texts or perform lengthy complex reasoning, \textsc{Visual-O1} did not significantly increase computational overhead and was more general and effective, even for general-intelligence models. We validated our approach across various tasks and models with different intelligence levels. Experimental results demonstrated that \textsc{Visual-O1} significantly improved the performance of models of varying intelligence on ambiguous instructions and enhanced their performance on general datasets. Our work revealed the potential of AI to operate like humans in real-world scenarios characterized by uncertainty and ambiguity.

For statements of broader impact, limitations, and reproducibility, please refer to the \textbf{Appendix} \ref{app:bi}, \ref{app:lim}, and \ref{app:rep}.

\bibliography{iclr2025_conference}
\bibliographystyle{iclr2025_conference}

\newpage

\appendix
\renewcommand{\thefigure}{\Alph{figure}}
\renewcommand{\thetable}{\Alph{table}}

This appendix mainly contains:
\begin{itemize}
    \item Supplementary implementation details in Section \ref{app:imp}
    \item Supplementary dataset details in Section \ref{app:data}
    \item Deeper analysis on improvement in Section \ref{app:improve}
    \item Further generalizability verification on intelligence in Section \ref{app:int}
    \item Further generalizability verification on various tasks in Section \ref{app:task}
    \item Additional ablation of components in Section \ref{app:abl}
    \item Additional ablation of reasoning in Section \ref{app:process}
    \item Computational overhead in Section \ref{app:comp}
    \item Extra cases in Section \ref{app:case}
    \item Statement of broader impact in Section \ref{app:bi}
    \item Statement of limitations in Section \ref{app:lim}
    \item Statement of reproducibility in Section \ref{app:rep}
\end{itemize}

\section{Supplementary Implementation Details}
\label{app:imp}

\textsc{Visual-O1} uses the following prompts for high-intelligent and general-intelligent models during inference. For each prompt, the upper part is used by the high-intelligent model, while the lower part is used by the general-intelligent model. 

\begin{tcolorbox}[title = {Prompt of Reasoning $p_\mathrm{rsn}$},
  fonttitle = \bfseries, fontupper = \ttfamily, colupper = black!50!white, fontlower = \ttfamily, collower = black!50!white, breakable]
You are a helpful assistant in normal conversation. \\
\textcolor{black}{\{task description\}} \\
Follow these instructions carefully: \\
1. Read the given question carefully and reset counter between \textbox[orange]{budget} \\
2. Generate a detailed, logical step-by-step solution. \\
3. Enclose each step of your solution within \textcolor{black}{ reasoning tags}. \\
4. You are allowed to use at most {budget} steps (starting budget), keep track of it by counting down within tags \textbox[orange]{budget}, STOP GENERATING MORE STEPS when hitting 0. You don't have to use all of them. \\
\\\\
Example format:\\\\
\textbox[orange]{starting budget}\\
\textcolor{black}{Content of step 1}\\\\
\textbox[orange]{remaining budget}\\
\textcolor{black}{Content of step 2}\\\\
\textbox[orange]{remaining budget}\\
\textcolor{black}{Content of step 3 or Content of some previous step}\\\\
\textbox[orange]{remaining budget}\\
...\\\\
\textbox[orange]{remaining budget}\\
\textcolor{black}{Content of final step}\\
\\\\
Description: \textcolor{black}{\{ambiguous instruction\}}\\
Provide a detailed, step-by-step solution to a given question.
\tcblower
\textcolor{black}{\{experience\}} \textcolor{black}{\{ambiguous instruction\}}
\end{tcolorbox}

Since the reasoning process of the high-intelligent model occurs during the inference stage, to accelerate this stage, we combine all the prompts, allowing the model to complete the entire reasoning process in a single output. To clearly demonstrate, we split the prompt used by the high-intelligent model into $p_\mathrm{rsn}$, $p_\mathrm{rfl}$, and $p_\mathrm{syn}$.

\begin{tcolorbox}[title = {Prompt of Reflection $p_\mathrm{rfl}$},
  fonttitle = \bfseries, fontupper = \ttfamily, colupper = black!50!white, fontlower = \ttfamily, collower = black!50!white, breakable]
Follow these instructions carefully: \\
5. Do a self-reflection when you are unsure about how to proceed; based on the self-reflection and reward, decide whether you need to return to the previous steps. \\
6. Provide a critical, honest, and subjective self-evaluation of your reasoning process within <reflection> and </reflection> tags. \\
7. Assign a quality score to your solution as a float between 0.0 (lowest quality) and 1.0 (highest quality), enclosed in <reward> and </reward> tags. \\
8. If the image or question is not clear enough, you need to reflect and try to get answers from the unclear image or question. \\
\\\\
Example format:\\
<reflection> [Evaluation of the solution] </reflection>\\
<reward> [Float between 0.0 and 1.0] </reward>\\
\tcblower
According to the instruction you generated last time, the annotator has rewritten \textcolor{black}{\{ambiguous instruction\}} as \textcolor{black}{\{clear instruction\}}. Please correct or rewrite your instruction based on the image situation. The image and the result of the data annotator are only for your evaluation. Please do not include the specific case in the instructions. You need to generate the full instruction even if no change is needed.
\\\\
If the annotator cannot find it, please let the annotator guess the one with the highest probability.
\\\\
Make sure the annotator only responds to the rewritten phrase and does not include any other thing.
\\\\
Instruction:
\textcolor{black}{\{experience\}}
\end{tcolorbox}

\begin{tcolorbox}[title = {Prompt of Response Synthesis $p_\mathrm{syn}$},
  fonttitle = \bfseries, fontupper = \ttfamily, colupper = black!50!white, fontlower = \ttfamily, collower = black!50!white, breakable]
After completing the solution steps, reorganize and synthesize the steps into the final answer within <answer> and </answer> tags. The final answer cannot be empty. 
\tcblower
Disambiguated question: \textcolor{black}{\{clear instruction\}}
\\
Original question: \textcolor{black}{\{ambiguous instruction\}}
\end{tcolorbox}

\section{Supplementary Dataset Details}
\label{app:data}

\begin{table}[t!]
\caption{\textbf{Distribution of dataset.}}
\label{tab:dis}
\begin{center}
\begin{small}
\begin{sc}
\begin{tabular}{ccccc}
\toprule
 {Ellipsis} &{Colloquialism} & {Subjectivity} & {Relativity} &{Other}\\
\midrule
$23.3\%$ & $27.3\%$ & $3.3\%$ & $29.3\%$ & $16.7\%$ \\
\bottomrule
\end{tabular}
\end{sc}
\end{small}
\end{center}
\end{table}

For RIS, we use the \textsc{RefCOCO+} dataset \citep{kazemzadeh2014referitgame}; for VQA, we use the \textsc{VizWiz} dataset \citep{gurari2018vizwiz}. For their ambiguous instructions, we manually screen and construct subsets of $150$, $650$, and $106$, respectively. Specifically, we use \textsc{GPT-4} for an initial screening, selecting $2,000$ sets of potentially ambiguous instructions from the original dataset. Then, we enlist multiple volunteers to manually screen for instructions that clearly contain ambiguity.

We categorize the ambiguous instructions into five distinct types based on the underlying causes of the ambiguity: \textit{ellipsis}, \textit{colloquialism}, \textit{subjectivity}, \textit{relativity}, and \textit{other}. The distribution of ambiguities can be found in Table \ref{tab:dis}. Their detailed explanations are as follows:

\begin{itemize}
    \item \textit{Ellipsis} indicates that the ambiguity stems from omitted content, where essential information is left out, leading to uncertainty about the intended meaning. 
    \item \textit{Colloquialism} refers to ambiguity arising from using informal or imprecise expressions that may not be universally understood or may vary in interpretation across different contexts.
    \item \textit{Subjectivity} indicates ambiguities due to subjective judgments, where personal opinions or individual perspectives cause unclear or varied interpretations.
    \item \textit{Relativity} indicates that ambiguity comes from implied comparisons, where the meaning depends on an unstated reference point or context, making the instruction open to multiple interpretations.
    \item \textit{Other} encompasses all other types of ambiguity that do not fit neatly into the previously mentioned categories, covering a broad range of miscellaneous sources of confusion.
\end{itemize}

We will release the data we set up under the MIT license.

\section{Deeper Analysis on Improvement}
\label{app:improve}

\begin{table}[t!]
\caption{\textbf{The improvements on different types of ambiguity.}}
\setlength{\tabcolsep}{0.4mm}
\label{tab:improve}
\begin{center}
\begin{small}
\begin{sc}
\begin{tabular}{lcccccccccccccccccccc}
\toprule
\multirow{2}{*}{Method} & \multicolumn{2}{c}{Ellipsis} & &  \multicolumn{2}{c}{Colloquialism} & &\multicolumn{2}{c}{Subjectivity} & &\multicolumn{2}{c}{Relativity} & &\multicolumn{2}{c}{Other}\\
\cmidrule{2-3}\cmidrule{5-6}\cmidrule{8-9}\cmidrule{11-12}\cmidrule{14-15} & gIoU & cIoU & & gIoU & cIoU& & gIoU & cIoU& & gIoU & cIoU& & gIoU & cIoU \\
\midrule
LISA & $0.0424$ & $0.0460$ && $0.0107$ & $0.0124$ && $0.0025$ & $0.0081$ && $0.0113$ & $0.0119$ && $0.0490$ & $0.0633$\\
Visual-O1  & $\mathbf{0.1742}$ & $\mathbf{0.1704}$ && $\mathbf{0.1662}$ & $\mathbf{0.1375}$  && $\mathbf{0.0826}$ & $\mathbf{0.2637}$  && $\mathbf{0.1444}$ & $\mathbf{0.1207}$  && $\mathbf{0.0763}$ & $\mathbf{0.0490}$\\
\bottomrule
\end{tabular}
\end{sc}
\end{small}
\end{center}
\end{table}

To deeply analyze how \textsc{Visual-O1} mitigates ambiguity issues, we separately calculate the original scores and the scores after using \textsc{Visual-O1} for each ambiguity category in RIS, as shown in Table \ref{tab:improve}. We find that, in the original case, LISA performs relatively best in \textit{ellipsis}, because \textit{ellipsis} is the relatively simplest category, while the scores for the other categories are almost all around $1$. After using Visual-O1, we observe significant improvements in scores across various categories, especially in \textit{colloquialism} and \textit{relativity}. This is because these two can be more easily converted into clear instructions through visual context. The significant improvements in other more challenging categories also signify the effectiveness of \textsc{Visual-O1}.

\section{Further Generalizability Verification on Intelligence}
\label{app:int}

\begin{table}[t!]
\caption{\textbf{General instruction results on different intelligence level.}}
\label{tab:int_ris_gen}
\begin{center}
\begin{small}
\begin{sc}
\begin{tabular}{lccc}
\toprule
Model & Visual-O1 & gIoU & cIoU \\
\midrule
LISA \citep{lai2023lisa} & &  $0.4654$ & $0.4721$ \\
Visual-O1 (LISA) & Empirical & $\mathbf{0.4794}$ & $\mathbf{0.5004}$ \\
Visual-O1$^+$ (LISA) & Instantial & $\mathbf{0.4985}$ & $\mathbf{0.5188}$ \\
SoM \citep{yang2023set}& & $0.3507$ & $0.4154$ \\
Visual-O1$^-$ (SoM) & Empirical & $\mathbf{0.3772}$ & $\mathbf{0.4336}$ \\
Visual-O1 (SoM) & Instantial & $\mathbf{0.3602}$ & $\mathbf{0.4566}$ \\
\bottomrule
\end{tabular}
\end{sc}
\end{small}
\end{center}
\end{table}

We not only compare the results on ambiguous instructions but also extend the comparison to general instructions to better confirm the generalization capability of \textsc{Visual-O1}. In Table \ref{tab:int_ris_gen}, we can also see that all variants of \textsc{Visual-O1} maintain and even improve performance on general instructions. This indicates that the design of \textsc{Visual-O1} is highly flexible and can be broadly applied to various intelligence levels of models, thereby improving the whole system's reasoning efficiency.

\section{Further Generalizability Verification on Extra Tasks}
\label{app:task}

\label{sec:exp_task}

\begin{table}[t!]
\caption{\textbf{Generalization studies on VLN.}}
\label{tab:gen_vln}
\begin{center}
\begin{small}
\begin{sc}
\begin{tabular}{lccc}
\toprule
Model & SR & SPL & Navi Error \\
\midrule
VLN-SIG \citep{li2023improving} & $4.72$ & $4.59$ & $7.95$\\
Visual-O1 (VLN-SIG) & $\mathbf{26.42}$ & $\mathbf{22.66}$ & $\mathbf{5.99}$\\
\bottomrule
\end{tabular}
\end{sc}
\end{small}
\end{center}
\end{table}

\begin{figure}[hpt!]
\vskip -0.2in
	\centering
	\includegraphics[width=13cm]{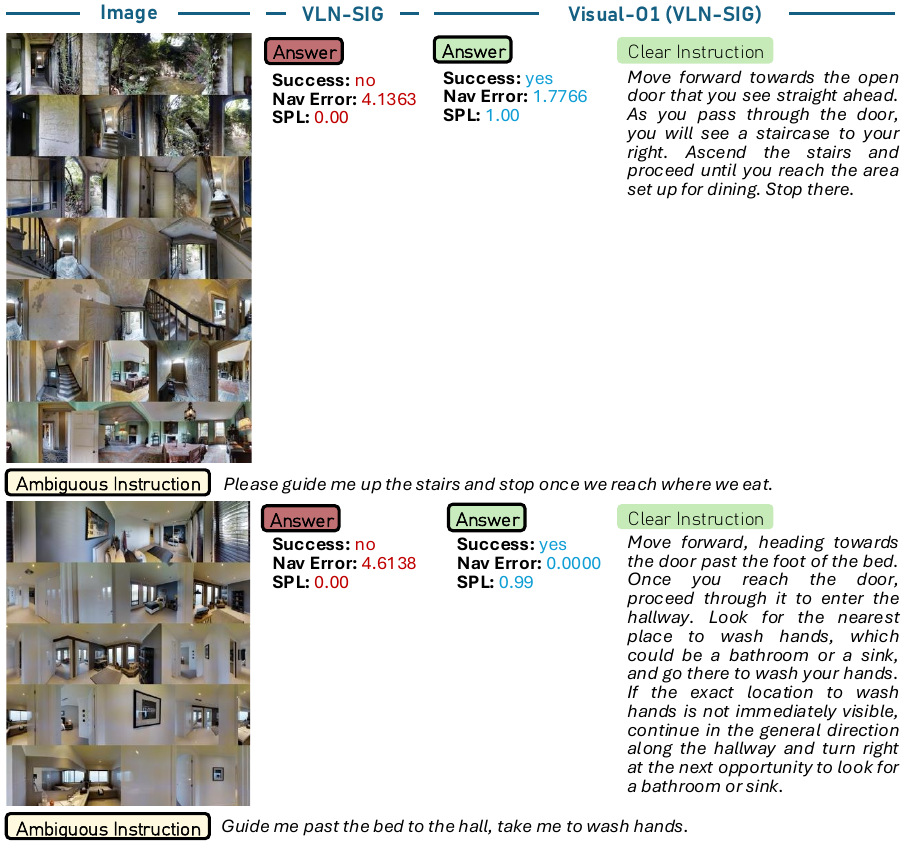}
	\caption{\textbf{Cases of Visual-O1 on VLN.}}
	\label{fig:vln}
\end{figure}

\begin{figure}[hpt!]
	\centering
 \vskip -0.2in
	\includegraphics[width=13cm]{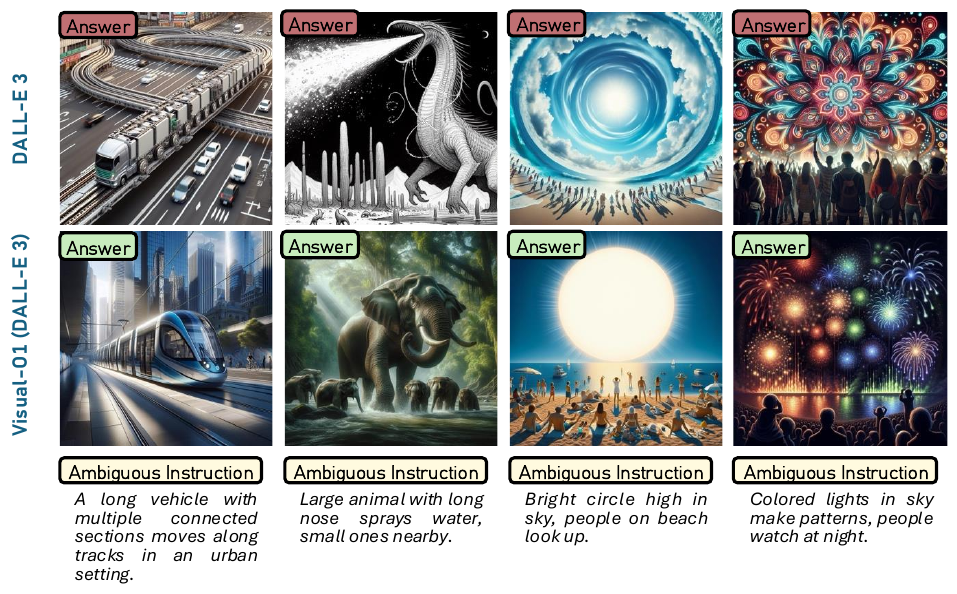}
	\caption{\textbf{Cases of Visual-O1 on Image Synthesis.}}
	\label{fig:ig}
\end{figure}

\textsc{Visual-O1} demonstrates remarkable generalizability across various tasks. To further validate this, we apply \textsc{Visual-O1} to two complex multi-modal tasks: image synthesis in the visual synthesis field and vision-and-language navigation (VLN) in the robotics field. Since models for these types of tasks do not have direct language output capabilities, we use \textsc{GPT-4o} for parts requiring intermediate language output. For VLN, we use the valid unseen split of the \textsc{Room-to-Room} dataset \citep{anderson2018vision}. As detailed in Table \ref{tab:gen_vln}, our observations reveal substantial performance enhancements across all tasks, underscoring the versatility and robustness of \textsc{Visual-O1}. We also provide examples in Figure \ref{fig:vln}. For image synthesis, we choose the state-of-the-art model \textsc{DALL-E 3}. As shown in Figure \ref{fig:ig}, we notice that even the most advanced models often misunderstand ambiguous instructions in human interactions, whereas \textsc{Visual-O1} significantly alleviates this issue. This broad applicability paves the way for extending our method to a wider array of applications, showcasing its potential to improve performance in diverse contexts.

\section{Additional Ablation of Components}
\label{app:abl}

\begin{table}[t!]
\caption{\textbf{Ablation studies.} }
\label{tab:abl2}
\begin{center}
\begin{small}
\begin{sc}
\begin{tabular}{lcc}
\toprule
Model & Acc & BLEU-1 \\
\midrule
Visual-O1 (LLaVA) & $\mathbf{57.38}$ & $\mathbf{0.7329}$ \\
\quad w/o Multi-examples & $55.75$ & $0.7092$ \\
\quad w/o Multi-modalities & $55.90$ & $0.7143$ \\
\bottomrule
\end{tabular}
\end{sc}
\end{small}
\end{center}
\end{table}

We further verify the effectiveness of \textsc{Visual-O1}'s one-time optimization design details in Table \ref{tab:abl2}. We find that without multiple examples, \textit{i.e.}, only allowing optimization for the same sample leads to overfitting and ineffective optimization. Without multiple modalities, \textit{i.e.}, lacking visual contexts, \textsc{Visual-O1}'s optimization capability also declines, as visual contexts help the optimizer determine the current state of optimization. By avoiding all these issues, it demonstrates excellent capabilities.

\section{Additional Ablation of Reasoning}
\label{app:process}

To better demonstrate \textsc{Visual-O1}'s reasoning process and reveal the specific content of \textsc{Visual-O1}'s reasoning, we show the detailed instantaneous and empirical experiences after each reasoning step of \textsc{Visual-O1}. The upper part is the instantaneous experience for the highly intelligent model, while the lower part is the empirical experience for the generally intelligent model.

\begin{tcolorbox}[title = {Experience Prompt $\mathcal{A}_\mathrm{ins}$ and $\mathcal{A}_\mathrm{emp}$},
  fonttitle = \bfseries, fontupper = \ttfamily, fontlower = \ttfamily, breakable]
  \begin{minipage}{0.3\textwidth}
    \includegraphics[width=\linewidth]{Figure/case.png}
  \end{minipage}
  \hfill
  \begin{minipage}{0.65\textwidth}
  between two bears and not plate
  \end{minipage}
  \\\\
\textbox[orange]{Budget 7} 
Identify the objects marked as bears in the image. 
\\\\
\textbox[orange]{Budget 6} 

The bears in the image are part of the design on the mugs, which are marked with [7] and [3].
\\\\
\textbox[orange]{Budget 5} 

Locate objects that are not plates. In this image, the plate is a prominent object, but we need to identify objects that are not plates.
\\\\
\textbox[orange]{Budget 4} 

Exclude the objects that are on the plate, such as the sandwich marked with [4], the soup bowl marked with [2], and the pickle marked with [8].
\\\\
\textbox[orange]{Budget 3} 

Identify any objects that are between the two bears but are not on the plate.
\\\\
\textbox[orange]{Budget 2} 

\textcolor{black!50!white}{It seems that the description provided was to find an object that is situated between the two bears but is not a plate itself. The steps taken so far have led to the exclusion of items on the plate and the identification of the bears on the mugs.
\\
<reward> 0.8 </reward>
}
\\\\
\textbox[orange]{Budget 1} 

Looking at the objects, the mugs marked with [7] and [3] are situated on either side of the image. The only object that is directly between them and not on the plate seems to be the salt or pepper shaker marked with [5].
\\\\
\textbox[orange]{Budget 0} 

The object that fits the description "between two bears and not plate" is the salt or pepper shaker marked with [5].

\textcolor{black!50!white}{The process involved a systematic exclusion of objects on the plate and identifying the bears on the mugs. The final object that fit the description was found to be between the bears (mugs) and not on the plate. The reasoning was sound, }
\textcolor{black!50!white}{and followed the instructions accurately.
\\
<reward> 1.0 </reward>
}
\tcblower
\textbox[orange]{Budget 3} Repeat the question.
\\\\
\textbox[orange]{Budget 2} Please review the image provided and use its details to rephrase the ambiguous and blurry question into a clear and precise one that can be effectively answered. Ensure that your revised question is directly related to the content of the image.
\\\\
\textbox[orange]{Budget 1} Please ensure that your revised question is specific and directly references observable elements in the image. Your question should guide the respondent in addressing particular details or aspects present in the image clearly despite any blurriness. Avoid general or vague terms and aim for specificity that will elicit a precise answer.
\\\\
\textbox[orange]{Budget 0} When rephrasing the question, focus on the discernible elements in the image, such as text, icons, or specific features visible on the computer monitor. Your question should ask for details about these specific elements, avoiding any reference to the clarity of the picture or the physical location, as these are not relevant to the content displayed on the screen. Aim to formulate a question that inquires about the information or processes shown in the image, which can be answered with the visible data.

\end{tcolorbox}

As the reasoning progresses, in the instantial experience, \textsc{Visual-O1} begins to perform increasingly in-depth reasoning. Meanwhile, in the empirical experience, \textsc{Visual-O1} starts to generate more specific requirements, such as generating specific locations and elements, which are essential for accurately interpreting and generating unambiguous instructions. This also proves the effectiveness of \textsc{Visual-O1}'s optimization.

\section{Computational Overhead}
\label{app:comp}

\begin{table}[t!]
\caption{\textbf{Computational overhead of Visual-O1.} }
\label{tab:comp}
\begin{center}
\begin{small}
\begin{sc}
\begin{tabular}{lccc}
\toprule
Method & Stage & Time & VRAM \\
\midrule
LLaVA & Lora Fine-tuning \citep{hu2021lora} & $25.4439$min & $88797$MB \\
Visual-O1 (LLaVA) & One-time Optimization & $1.5260$min & $16148$MB \\
\midrule
LLaVA & Inference & $0.5103$s & $16102$MB \\
Visual-O1 (LLaVA) & Inference & $0.6547$s & $16370$MB \\
\midrule
GPT-4o & Inference & $4.9403$s & - \\
Visual-O1 (GPT-4o) & Inference & $6.7614$s & - \\
\bottomrule
\end{tabular}
\end{sc}
\end{small}
\end{center}
\end{table}

As shown in Table \ref{tab:comp}, we analyze the computational overhead of \textsc{Visual-O1}. We find that \textsc{Visual-O1}'s overhead in both the optimization and disambiguation phases is low. Specifically, we also compare the time it takes for the vanilla \textsc{LLaVA} and \textsc{GPT-4o} to complete a VQA task. The computational overhead is comparable to \textsc{LLaVA}, further proving \textsc{Visual-O1}'s capability.

\section{Extra Cases}
\label{app:case}

\begin{figure}[hpt!]
	\centering
	\includegraphics[width=13cm]{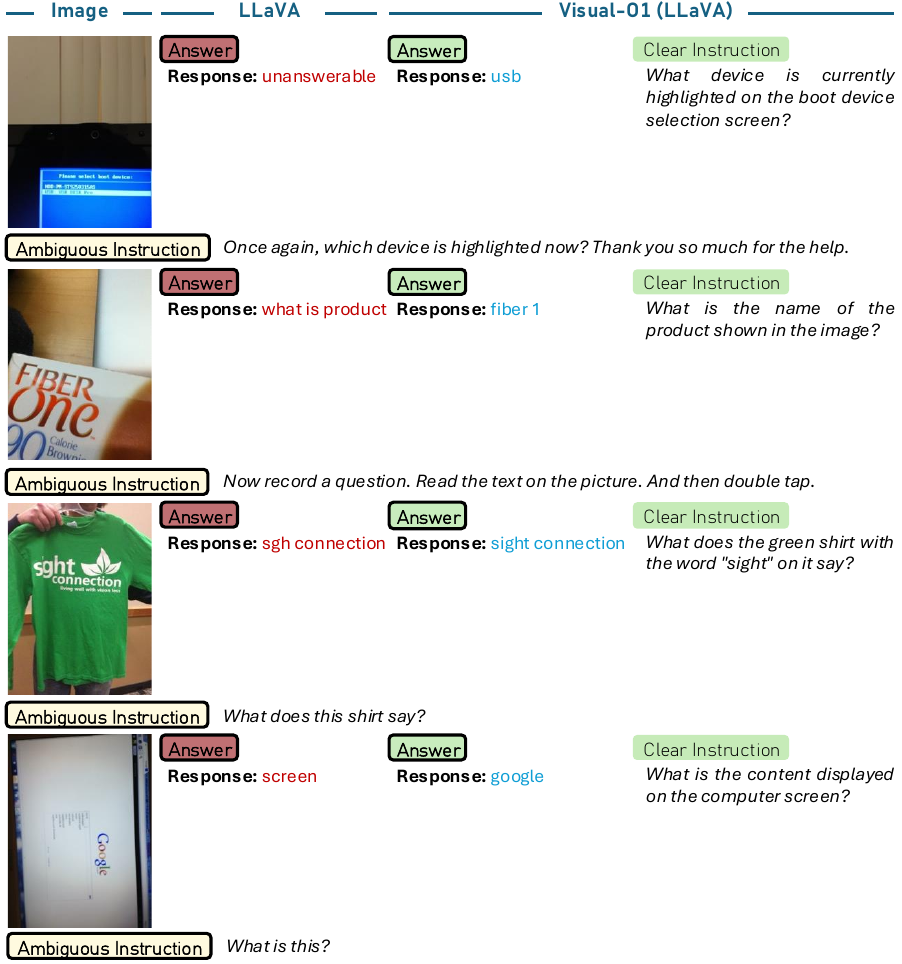}
	\caption{\textbf{Case studies on VQA.}}
	\label{fig:vqa}
\end{figure}

Similar to RIS, we also visualize examples on VQA. As shown in Figure \ref{fig:vqa}, \textsc{Visual-O1} significantly improves the clarity of ambiguous instructions, aiding the task model in achieving correct and natural results.

\section{Broader Impact}
\label{app:bi}

With the development and application of AI, language instructions for interacting with AI are being applied in an increasing number of scenarios. In reality, humans often issue vague instructions for communication, and due to their natural visual ability, language and visual information complement each other, which further intensifies the ambiguity of instruction information. In this paper, we reveal the phenomenon of ambiguous instructions. Our proposed \textsc{Visual-O1} significantly alleviates the pressure of ambiguous instructions on AI models, helps AI understand more natural instructions, broadens the application range of AI, and serves more non-professional people.

\section{Limitations}
\label{app:lim}

Despite \textsc{Visual-O1} demonstrating strong reasoning capabilities and allowing general-intelligent models like \textsc{LLaVA} to perform inference without relying on large-scale high-intelligent models such as \textsc{GPT-4o}, it still requires the use of high-intelligent models during the one-time optimization phase. In the future, we will explore methods that do not rely on high-intelligent models even during optimization, in order to better extend our approach to scenarios with more limited resources.

\section{Reproducibility Statement}
\label{app:rep}

We place a high emphasis on the reproducibility of our work. To facilitate this, we have provided a comprehensive set of implementation details and prompts in the appendix. Additionally, to further enhance the reproducibility of our results, we will release the source code and data. 

\end{document}